\documentclass{article}

\usepackage{microtype}
\usepackage{graphicx}
\usepackage{amsfonts}
\usepackage{subfigure}
\usepackage{amstext}
\usepackage{booktabs} % for professional tables
\usepackage{amsmath}

\usepackage{hyperref}

\usepackage[accepted]{icml2019}

\icmltitlerunning{Sample-efficient Adversarial Imitation Learning from Observation}

\begin{document}
\twocolumn[
\icmltitle{Sample-efficient Adversarial Imitation Learning from Observation}



\icmlsetsymbol{equal}{*}

\begin{icmlauthorlist}
\icmlauthor{Faraz Torabi}{ut}
\icmlauthor{Sean Geiger}{ut}
\icmlauthor{Garrett Warnell}{army}
\icmlauthor{Peter Stone}{ut}
\end{icmlauthorlist}

\icmlaffiliation{ut}{University of Texas at Austin, Austin, USA}
\icmlaffiliation{army}{U.S. Army Research Laboratory}

\icmlcorrespondingauthor{Faraz Torabi}{faraztrb@cs.utexas.edu}

\icmlkeywords{Machine Learning, ICML, Imitation Learning}

\vskip 0.3in
]



\printAffiliationsAndNotice{} % otherwise use the standard text.

\begin{abstract}
%Recently, sample efficient trajectory centric reinforcement learning algorithms have been successfully combined with neural network function approximations which resulted in impressive performance in learning complex behaviors directly on physical robots. 
Imitation from observation is the framework of learning tasks by observing demonstrated state-only trajectories. Recently, adversarial approaches have achieved significant performance improvements over other methods for imitating complex behaviors. 
However, these adversarial imitation algorithms often require many demonstration examples and learning iterations to produce a policy that is successful at imitating a demonstrator's behavior.
This high sample complexity often prohibits these algorithms from being deployed on physical robots.
In this paper, we propose an algorithm that addresses the sample inefficiency problem by utilizing ideas from trajectory centric reinforcement learning algorithms.
We test our algorithm and conduct experiments using an imitation task on a physical robot arm and its simulated version in Gazebo and will show the improvement in learning rate and efficiency.
%We find that this new method results in successful imitation learning with fewer samples than generative adversarial imitation networks alone.
\end{abstract}

\section{Introduction}
\label{Introduction}

Teaching new actions to robot actors through demonstration is one of the most attractive methods for behavior learning. 
While robots can learn new behaviors using reinforcement learning with a pre-specified reward function \cite{Sutton1998}, significant exploration is often required to extract the behavior from the reward. 
In some cases, denser reward functions can help speed up the exploration process, but designing them requires a certain level of skill and understanding of the reinforcement learning process, and can often result in unexpected behaviors when the reward function doesn\textquoteright t precisely guide the action.
Instead, teaching a robot a behavior simply by demonstrating it removes the requirement of explicitly specifying a reward function altogether.
Anyone who knows how to perform the task can demonstrate it without understanding the learning process, and the learning process requires much less exploration. 
This process\textendash learning from demonstration (LfD)\textendash aims to take a series of observed states (e.g. joint angles, position in space) and actions (e.g. decisions to move a joint at some speed) and extract a policy that approximates the demonstrated behavior \cite{Argall2009}.

While being able to imitate a behavior after observing the state and actions of a demonstrator is useful, there are many situations where the actions of the demonstrator are unknown. 
Common approaches to LfD require both the states and actions of the demonstrator to be recorded \cite{Argall2009}. 
In imitation from external observation (IfO) \cite{liu2018imitation,torabi2019recent}, on the other hand, just the observable states of the demonstrator are known-{}-no action information is available.  
Imitating behaviors solely from observable data greatly expands the set of possible demonstrators: behaviors could be learned from in-person human demonstrators or even the vast collection of videos available online.

While imitation from external observation has been studied and performed with some success for two decades \cite{ijspeert2001trajectory}, recent advances in deep neural networks have widened the set of behaviors that can be imitated and the ways that demonstration data can be collected.
One way deep learning has been applied to IfO is through generative adversarial networks \cite{torabi2018generative,ho2016generative,chen2016infogan}. 
In this approach-{}-generative adversarial imitation from observation (GAIfO)-{}-one network learns a control policy for imitating the demonstrator while the other learns to discriminate between the demonstrator's behavior and that of the imitator. 
While GAIfO advanced the state of the art in imitation from observation, it comes with its own set of challenges. 
First, in comparison with simpler regressed models, deep networks are notorious for requiring orders of magnitude more training data, and GAIfO is no exception. 
Second, this algorithm uses model-free reinforcement algorithms which are usually very data inefficient.
Some of the possible benefits of the applications of IfO break down when a high sample size is required.
Therefore, in practice, this algorithm has been largely limited to being studied in simulation. 
In simulation, many experiences and large demonstration sets can be collected quickly. 
Physical demonstrations are more costly to perform, and real-time constraints limit the speed at which control policies can be evaluated and thus behavior learned. 
%The sample size required to train the adversarial networks places a limit on the practicality of collecting in real-time on a real robot. 
For imitation from observation to work on a physical robot, a higher degree of sample efficiency is required.

Deep reinforcement learning has faced similar obstacles with learning with limited samples, especially in the context of robotic control policies with complex dynamics. 
However, recently, trajectory centric reinforcement learning algorithms are being used to guide neural network policy search which has been shown that is very sample-efficient \cite{Levine2013,levine2014learning,Levine2015a,levine2016end}.
These algorithms achieve this sample efficiency in part by gaining insight into dynamics through the iterative training of linear quadratic regulators (iLQR's) \cite{tassa2012synthesis} on a set of trajectory controllers.

In this paper, we propose an imitation from observation algorithm, LQR+GAIfO, that takes advantage of both (1) the high performance of the adversarial learning algorithms, and (2) the sample efficiency of trajectory centric reinforcement learning algorithms.
%seek to answer the question:\emph{ will using the techniques
%of linear quadratic regulators in combination with generative adversarial
%imitation from observation provide a more sample-efficient approach
%to imitation learning? }To investigate this question, we propose an
%algorithm that combines GAIfO and LQR and hypothesize that this technique
%will allow an imitator to approximate a demonstrator's actions in
%fewer iterations than GAIfO alone. 
We apply the proposed algorithm to a 6-degree-of-freedom robot arm to learn to imitate behaviors from a set of low-level state trajectories. We find that this new method results in successful imitation learning with fewer samples than the previous algorithms.

In Section \ref{sec:related-work} of this paper, we discuss previous work related to this
topic. In Section \ref{sec:prem}, we cover the techniques involved in GAIfO and
LQR. Section \ref{sec:alg}, describes our approach to combining LQR and GAIfO into
one functional algorithm. In Section \ref{sec:exp}, we share our experimental
setup and results, and we discuss results in Section \ref{sec:disc}. Finally, in
Section \ref{sec:conclusion}, we summarize and discuss potential future work.

\section{Related Work}\label{sec:related-work}

Our approach to sample-efficient imitation learning is built upon previous works in the field of imitation learning and trajectory-centric reinforcement learning. In the following we discuss previous works on both topics.

%as well as trajectory centric reinforcement learning.
%\subsection{Imitation Learning}
Techniques for imitation learning differ in the way they approach the problem. 
Two popular approaches to imitation learning have been behavioral cloning \cite{Pomerleau1991} and inverse reinforcement learning (IRL) \cite{ng2000algorithms,russell1998learning}.
Behavioral cloning views the imitation learning problem as a supervised learning problem that attempts to learn a direct mapping from states to actions. On the other hand, inverse reinforcement learning works to find a cost function under which the expert demonstrator is optimal. One approach of this type is guided cost learning \cite{finn2016guided} which builds on maximum entropy IRL \cite{ziebart2008maximum} and guided policy search algorithm \cite{levine2014learning} and achieves impressive results on physical robots.
Later, in \citeyear{ho2016generative}, \citeauthor{ho2016generative} used generative adversarial networks to imitate policies when both states and actions are available using a technique called generative adversarial imitation learning (GAIL) \cite{ho2016generative}.
One imitator network attempts to imitate the policy while another attempts to discriminate between the imitation and provided demonstration data\cite{goodfellow2014generative}. Several follow-up works have improved upon this approach on different aspects \cite{fu2018learning,song2018multi} and recently, there has been efforts to address sample efficiency of this algorithm by proposing approaches for unbiasing rewards and deriving an off-policy formulation of adversarial imitation learning algorithms \cite{kostrikov2019addressing}.
%Both networks are randomly initialized, and on each iteration, the policy of the imitator is executed, the discriminator is updated based on a loss function against the demonstration data, then the imitator is updated based on the loss of the discriminator.

These approaches however, require access to the demonstrator's actions. Recently, on the other hand, imitation learning from observation \cite{torabi2018behavioral,torabi2019recent} is becoming more popular in which the agent only has access to state demonstrations of the expert. 
An algorithm of this type is generative adversarial learning from observation (GAIfO) \cite{torabi2018generative,torabi2019adversarial,torabi2019imitation,stadie2017third} which uses a GANs like architecture to bring the state-transition distribution of the imitator closer to that of the demonstrator.
%More specifically, GAIL characterizes the cost function as a function of states and actions while the cost function in GAIfO is instead characterized by state transitions.
%GAIfO provides two implementations: one in that uses low-level state data (e.g., direct measurements of robot arm joint angles) to form a policy and another that handles imitation from visual data by adding convolutional layers at the front of the imitator and discriminator that take multiple frames of video as input. 
While this technique has been shown to discover accurate imitation policies, to date, they have only been evaluated in a simulated experimental domain. 
Because experiments consist of thousands of iterations in which each iteration includes executing a policy several times, the time required for monitoring experiments is prohibitive. 
%\subsection{Trajectory Centric Reinforcement Learning}
%

On the other hand, in reinforcement learning\textendash policy learning through environment-provided
reward functions only\textendash direct policy search in a large state-action
space requires numerous samples and often can fall into poor local
optima. Guided policy search (GPS) is a method to improve the sample
efficiency of direct policy search and guide learning in a large space
away from poor local optima \cite{Levine2013}. The basis of GPS is
to use trajectory optimization to focus policy learning on high-reward
actions. 

In guided policy search under unknown dynamics, time-varying linear
Gaussian models of the dynamics for a small set of specific tasks
are first trained to fit a small set of sample data through LQR \cite{levine2014learning}.
These Gaussian controllers are then sampled to generate samples to
optimize a general policy for a model with thousands of parameters
that would typically require much more training data. Specifically,
samples in regions of trajectories that have been found to lead to
higher reward are generated, guiding the policy learning.

GPS has had success in learning policies in reinforcement learning
situations with complex dynamics and high-dimensional inputs, including
training a policy that directly controls the torque on motors in a
robot arm to perform a task like screwing a cap on a bottle solely
from raw images of the system \cite{levine2016end}. Current applications
of GPS have focused on reinforcement learning, and the technique's
applications to IfO have not been adequately explored.

In this work, our goal is to resolve GAIfO’s sample inefficiency with the help of linear quadratic regulators to the extent that it can be applied to learning a behavior on a real robot.

\section{Preliminaries and Overview}\label{sec:prem}
In this section, we describe the notation considered throughout the paper, and the two methods that our proposed algorithm are based on, (1) adversarial imitation from observation, and (2) trajectory centric reinforcement learning.
%Our approach to sample-efficient imitation learning is based on work
%in two areas: generative adversarial imitation learning, which uses
%a dual-network approach to fitting a policy to expert data, and linear
%quadratic regulators, a sample-efficient approach to extracting a
%linear policy from a quadratic cost function. In this section, we
%review both of these techniques in detail.

\subsection{Notation}

We consider agents acting within the broad framework of Markov decision processes (MDPs). We denote a MDP using the 5-tuple $M=\{S, A, P, r, \gamma \}$, where $S$ is the agent's state space, $A$ is its action space, $P(s_{t+1}|s_t,a_t)$ is a function denoting the probability of the agent transitioning from state $s_t$ to $s_{t+1}$ after taking action $a_t$, $r: S \times A \rightarrow \mathbb{R}$ is a function specifying the immediate reward that the agent receives for taking a specific action in a given state, and $\gamma$ is a discount factor.
	In this framework, agent behavior can be specified by a policy, $\pi: S \rightarrow A$, which specifies the action (or distribution over actions) that the agent should use when in a particular state.
%	We denote the set of state transitions experienced by an agent during a particular execution of a policy ${\pi}$ by $\tau_{\pi} = \{(s_i, s_{i+1})\}$.
	
	In reinforcement Learning the goal is to learn a policy, $\pi$, by maximizing the accumulated reward, $r$, through interaction with the environment. However, imitation learning considers the setting of M\textbackslash r, i.e. the reward function is excluded. Instead the agent has access to some demonstrated trajectories. The problem that we are interested in this paper is imitation from observation where these demonstrations only include state-trajectories of the expert $\tau_{E} = \{s_t\}$.

%\subsection{Generative Adversarial Imitation Learning}

%Generative adversarial learning consists of training two networks
%that operate in tandem: a generative network $G$, that attempts to
%generate data similar to training data, and a discriminative network
%$D$, that learns to differentiate the generated data and the training
%data \cite{goodfellow2014generative}. $G$ is trained through backpropagation
%to increase the likelihood of $D$ making a mistake. At the unique
%optimal solution, $G$ perfectly fits the training data distribution,
%and $D$ cannot differentiate between the output of $G$ and the training
%data.
\subsection{Adversarial Imitation from Observation}\label{AIO)}

Generative adversarial imitation from observation \cite{torabi2018generative} is an algorithm of this type in which attempts to learn tasks by bringing the state transition distribution of the imitator closer to that of the demonstrator.
The algorithm works as follows. 
There is an imitator policy network, $\pi_{\phi}$, that is initialized randomly. This policy is then executed in the environment to generate trajectories $\tau_{\pi}$ where each
trajectory is a set of states $\{(s_0,s_1,...,s_{n})\}$.
There is also a discriminator network parameterized by weights $\theta$ and maps
input trajectories to a score between $0$ and $1$: $D_\theta:S\times A\rightarrow[0,1]$,
The discriminator is trained in a way to output values close to zero for the data coming from the expert and close to one for the data coming from the imitator. To do so, $\theta$ is updated by taking an iteration towards solving the following optimization problem.
\begin{equation}
\displaystyle\max_{D_\theta \in (0,1)^{\mathcal{S} \times\mathcal{S}}} \mathbb{E}_{\tau_\pi} [\log(D_\theta(s,s'))]+ \mathbb{E}_{\tau_E} [\log(1-D_\theta(s,s'))]
\end{equation}

%$$-\Big(\mathbb{E}_\tau [\log(D_\theta(s,s'))]+\mathbb{E}_{\tau_E} [\log(1-D_\theta(s,s'))]\Big)$$
%\begin{equation}
%\mathbb{\widehat{E}_{\tau_{\mathrm{I}}}}\left[\nabla_{w}log\left(D_{w}\left(\tau\right)\right)\right]+\mathbb{\widehat{E}_{\tau_{\mathrm{E}}}}\left[\nabla_{w}log\left(1-D_{w}\left(\tau\right)\right)\right]
%\end{equation}

%With an output value of $0$ denoting that the input trajectory matches
%the distribution of expert (demonstration) trajectories $\tau_{E}$,
%and an output of $1$ meaning the two distributions are separate. 
%The generative network is a policy parameterized by weights $\phi$
%that chooses actions when provided with states in order to produce
%a new trajectory. 
From a reinforcement learning point of view, the discriminator
network provides a cost function that could change $\phi$ to move the
distribution of trajectories created by $\pi_\phi$ towards the
distribution of the demonstrated trajectories $\tau_E$. 

%In each iteration of GAIL, a number of trajectories $\tau_{I}$ are
%sampled from the current imitation policy $\pi_{_{I}}$. These trajectories
%are used to approximate the gradients of their respective policies
%with respect to the output of $D$, with the approximate expected
%value denoted as $\mathbb{\widehat{E}_{\tau}}$. The weights $w$
%of the discriminator network are then updated according to the \emph{Adam
%} \cite{kingma2014adam} technique for optimization of objective functions
%along the gradient, which is given by:
%
%\begin{equation}
%\mathbb{\widehat{E}_{\tau_{\mathrm{I}}}}\left[\nabla_{w}log\left(D_{w}\left(\tau\right)\right)\right]+\mathbb{\widehat{E}_{\tau_{\mathrm{E}}}}\left[\nabla_{w}log\left(1-D_{w}\left(\tau\right)\right)\right]
%\end{equation}

Therefore, following the update to $D_\theta$, the imitator
policy, $\pi_\phi$, is updated using the technique of Trust Region Policy Optimization
 \cite{schulman2015trust} under the cost function
 \begin{equation}
 \log(D_\theta(s,s'))
 \end{equation}
where $D_{\theta}$ is the newly updated discriminator network. The whole process is repeated until convergence.

It is a quite well-known fact that model-free reinforcement learning algorithms (e.g. TRPO) often require a large number of environment interactions. Therefore, it is not practical to deploy these types of algorithms on physical robots. On the other hand, model-based RL algorithms have shown promising performance in the real world \cite{Levine2015a,levine2016end}.

%In generative adversarial imitation from observation \cite{torabi2018generative},
%actions are not available in the demonstration data and are not present
%in collected trajectories. The update steps remain the same, but the
%trajectories now only consist of states: $\tau=\{s_{0},s_{1},...,s_{n}\}$.

\subsection{Trajectory Centric Reinforcement Learning}%Guided Policy Search and Linear Quadratic Regulators}

Linear quadratic regulators (LQR's) learn control policies under two
assumptions \cite{Bemporad2002}:
\begin{enumerate}
\item The dynamics of the environment are linear. This means that the transition
from a particular state given an action $f(s_{t},a_{t})$ can be represented
as the product of the state/action and a matrix $F_{t}$ plus a constant
vector $f_{t}$:
\[
f(s_{t},a_{t})=F_{t}\left[\begin{array}{c}
s_{t}\\
a_{t}
\end{array}\right]+f_{t}
\]
\item The cost is quadratic. The cost is represented by a quadratic term
$C_{t}$ and a linear vector $c_{t}$:
\[
c(s_{t},a_{t})=\frac{1}{2}\left[\begin{array}{c}
s_{t}\\
a_{t}
\end{array}\right]^{T}C_{t}\left[\begin{array}{c}
s_{t}\\
a_{t}
\end{array}\right]+\left[\begin{array}{c}
s_{t}\\
a_{t}
\end{array}\right]^{T}c_{t}
\]
\end{enumerate}
 The algorithm attempts to solve an optimization problem that returns the actions that have the highest return in the course of an episode. 
  Solving this optimization problem, results in a linear controller:
  \begin{align}
  a_t = K_t s_t + k_t
  \end{align}
  where the $K_t$s and $k_t$s are matrices and vectors which are combinations of $F_t$s, $C_t$s, $f_t$s, and $c_t$s that can be computed for each time-step. 
 
 In situations where the
dynamics are assumed to be close to linear but are not completely
known or are non-deterministic, the linear transition function is often
replaced by a conditional probability specified under a normal Gaussian
distribution, with a mean of the linear dynamics and a covariance:
\[p(s_{t+1}|s_{t},a_{t})=\text{\ensuremath{\mathcal{N}}}(F_{t}\left[\begin{array}{c}
s_{t}\\
a_{t}
\end{array}\right]+f_{t},\sigma^{2})
\]

When the covariance is constant (independent of the state and action),
the optimal policy is identical to the non-stochastic LQR.

In non-linear systems where the cost is not quadratic, the techniques
of LQR can be used by approximating the dynamics with a first-order
Taylor expansion and approximating the cost with a second-order Taylor
expansion:
\[
F_{t}=\nabla_{s_{t},a_{t}}f(s_{t},a_{t}),\quad C_{t}=\nabla_{s_{t},a_{t}}^{2}c(s_{t},a_{t}),
\]
\[
\quad c_{t}=\nabla_{s_{t},a_{t}}c(s_{t},a_{t})
\]

Iterative linear quadratic regulators (iLQR's) can be used to find
optimal controllers under non-linear models by running LQR with the approximated
dynamics, then updating the dynamics fit on each iteration \cite{li2004iterative}. The resulting controller is:
$$a_t = K_t(s_t - \hat{s}_t)+k_t+\hat{a}_t$$
Where $\hat{s}_t$ and $\hat{a}_t$ are the states and actions around which the Taylor expansion is computed.

LQR assumes that the dynamics of the environment are known. Learning
dynamics for a given situation involves building a model to define
$f(s_{t},a_{t})$ from a set of observed state/action transitions
$\tau=\{(s_{t},a_{t},s_{t+1})\}$. A simple approach to this model building
is to use linear regression to estimate the dynamics, finding some
matrices $X$ and $Y$ that model the transition as $f(s_{t},a_{t})=Xs_{t}+Ya_{t}+c$,
or in a stochastic environment, $p(s_{t+1}|s_{t},a_{t})=\text{\ensuremath{\mathcal{N}}}(Xs_{t}+Ya_{t}+c,\sigma^{2})$.
Modelling dynamics with a Gaussian approximation of the linear regression
(often called linear Gaussian models) has the advantage of being very
sample-efficient.

To avoid the erroneous pursuit of an incorrect global optimal, a set
of local models can be used to replace a global model. The most expressive
case of local models is a set of models with a single model for every
time-step. In the linear regression approach, this amounts to fitting
new $X_{t}$ and $Y_{t}$ for every time-step, often called time-varying
controllers. Because dynamics are often highly correlated between
time-steps, this approach can be refined by using a global model as
a prior for a Bayesian linear regression at each time-step. For a better approximation of the local models it is shown that linear-Gaussian controllers, $p(a_t|s_t) = \mathcal{N} (K_t(s_t-\hat{s}_t)+k_t+\hat{a}_t, \Sigma_t)$, should be used for generating the training data \cite{levine2016end}. The covariance depends on the sensitivity of the total cost to the choice of action.

Because linear regression can overshoot optimals of non-linear dynamics,
policy adjustment can be bounded so that each iteration's update to
the model's transition distribution (or trajectory distribution) is
not too large. This can be achieved with a bound on the Kullback\textendash Leibler
(KL) divergence\textendash a relative measure of divergence between
distributions\textendash between the previous trajectory distribution
and the current trajectory distribution.

\section{Proposed Algorithm}\label{sec:alg}

In this section, we propose an imitation from observation algorithm, LQR+GAIfO, to learn an imitation policy from state only demonstrations, $\tau_{E} $. Our algorithm takes advantage of the high performance of adversarial imitation from observation algorithms and the sample efficiency of trajectory-centric reinforcement learning algorithms. To do so, we build upon the methods described in Section \ref{sec:prem}. For LQR to be useful in an imitation learning scenario, it can no longer depend on a pre-specified reward function that defines the task. Instead, the trajectory optimization step in LQR should be based on the existing controller's ability to imitate the expert demonstration. To achieve this capability, we train a discriminator network on each iteration and use an approximate version of its loss on the sampled trajectories to optimize the controllers.

Our algorithm begins by initializing the linear Gaussian controller and executing it inside the environment to collect state-action trajectories $\{(s_t, a_t)\}$. Then it randomly initializes a time-varying model $p$ to model the trajectory dynamics. $p$ is specified as $p(s_{t+1}|s_{t},a_{t})=\text{\ensuremath{\mathcal{N}}}(F_{t}\left[\begin{array}{c}
s_{t}\\
a_{t}
\end{array}\right]+f_{t},\sigma^{2})$. Given a set of state-action trajectories $\{s_t,a_t\}$,
$F_{t}$, $f_{t}$, and $\sigma^{2}$ are fit to the sample data at
each time-step using Bayesian linear regression with a normal-inverse-Wishart
prior. For this prior, it fits the entire trajectory sample to a Gaussian
mixture model (GMM), which previous research has found to
be effective \cite{levine2016end}. 
% every iteration of the algorithm, the current controller is used to collect $\left\{ \tau_{s}\right\} $, which is then used to refit the dynamics. 

\begin{algorithm}[t!]
	\caption{LQR+GAIfO}\label{alg:robotics}
	\begin{algorithmic}[1]
%		\STATE Initialize parametric policy $\pi_\phi$ with random $\phi$ 
		\STATE Initialize controller $p(a|s)$
		\STATE Initialize a neural network discriminator $D_\theta$ with random parameter $\theta$
		\STATE Obtain state-only expert demonstration trajectories $\tau_E=\{s_t\}$
		\WHILE {Controller Improves}
			\STATE Execute the controller, $p(a|s)$, and store the resulting trajectories $\tau_{p(a|s)}=\{(s,a,s')\}$
			\STATE Learn dynamics model $p(s'|s,a)$ over $\tau$
			\STATE Update $D_\theta$ using loss $$\displaystyle\min_{D_\theta^{\mathcal{S} \times\mathcal{S}}} \mathbb{E}_{\tau_{p(a|s)}} [D_\theta(s,s')]- \mathbb{E}_{\tau_E} [D_\theta(s,s'))]$$
			\STATE Create the composite function $C(s_{t},a_{t}) = (D_\theta \circ f_t)(s_t, a_t)$
			\STATE Compute the quadratically approximated cost function by taking the second order Taylor expansion of $C(s_{t},a_{t})$
% the  $c(s,a)$ by solving
%$$\displaystyle{\argmin_{C_t,c_t}}~\mathbb{E}_{\tau_\pi} [||c(s,a)+D_\theta(o,o')||^2]$$
\[
c_{q}(s_{t},a_{t})=\frac{1}{2}\left[\begin{array}{c}
s_{t}\\
a_{t}
\end{array}\right]^{T}\nabla_{s,a}^{2}C(s_{t},a_{t})\left[\begin{array}{c}
s_{t}\\
a_{t}
\end{array}\right]+
\]
\[
\left[\begin{array}{c}
s_{t}\\
a_{t}
\end{array}\right]^{T}\nabla_{s,a}C(s_{t},a_{t})
\]
%			\STATE Execute controller $p(a|s)$ and store the resulting trajectories $\tau = (s,o,a,s',o')$
%			\STATE Learn a neural network policy $\pi_\phi(a|o)$ by a supervised learning algorithm over $\tau$
			\STATE Improve controller $p(a|s)$ by LQR
		\ENDWHILE
	\end{algorithmic}
\end{algorithm}	

Following the dynamics model update, a randomly initialized neural network is considered as the discriminator, $D_\theta$, which takes state-transitions $(s_t, s_{t+1})$ as input and outputs a value. Similar to Section \ref{AIO)}, The goal is to train the discriminator to distinguish between the state-transitions coming from the controller and the demonstrator. However, in order to stabilize the learning, our algorithm uses Wasserstein loss \cite{arjovsky2017wasserstein} and takes an iteration on the following optimization problem.
%\begin{equation}
$$\displaystyle\min_{D_\theta^{\mathcal{S} \times\mathcal{S}}} \mathbb{E}_{p(a|s)} [D_\theta(s,s')]- \mathbb{E}_{\tau_E} [D_\theta(s,s'))]$$
%\end{equation}
Gradient penalties are also used as the regularization for further stabilization of the learning process \cite{gulrajani2017improved}. As discussed in Section \ref{sec:prem}, the discriminator\textemdash a function of state-transition $(s_t, s_{t+1})$\textemdash can be used as the cost function for training the controller. However, LQR requires the cost function to be a quadratic function of states and actions. Therefore, first, the discriminator is combined with the Gaussian
dynamics models to create a composite cost function $C(s_{t},a_{t}) = (D_\theta \circ f_t)(s_t, a_t)$. This composite
function is then quadratically approximated by taking the second
order Taylor expansions of the cost:

\[
c_{q}(s_{t},a_{t})=\frac{1}{2}\left[\begin{array}{c}
s_{t}\\
a_{t}
\end{array}\right]^{T}\nabla_{s,a}^{2}C(s_{t},a_{t})\left[\begin{array}{c}
s_{t}\\
a_{t}
\end{array}\right]+
\]
\[
\left[\begin{array}{c}
s_{t}\\
a_{t}
\end{array}\right]^{T}\nabla_{s,a}C(s_{t},a_{t})
\]

\begin{figure}
\begin{centering}
\includegraphics[width=0.35\paperwidth]{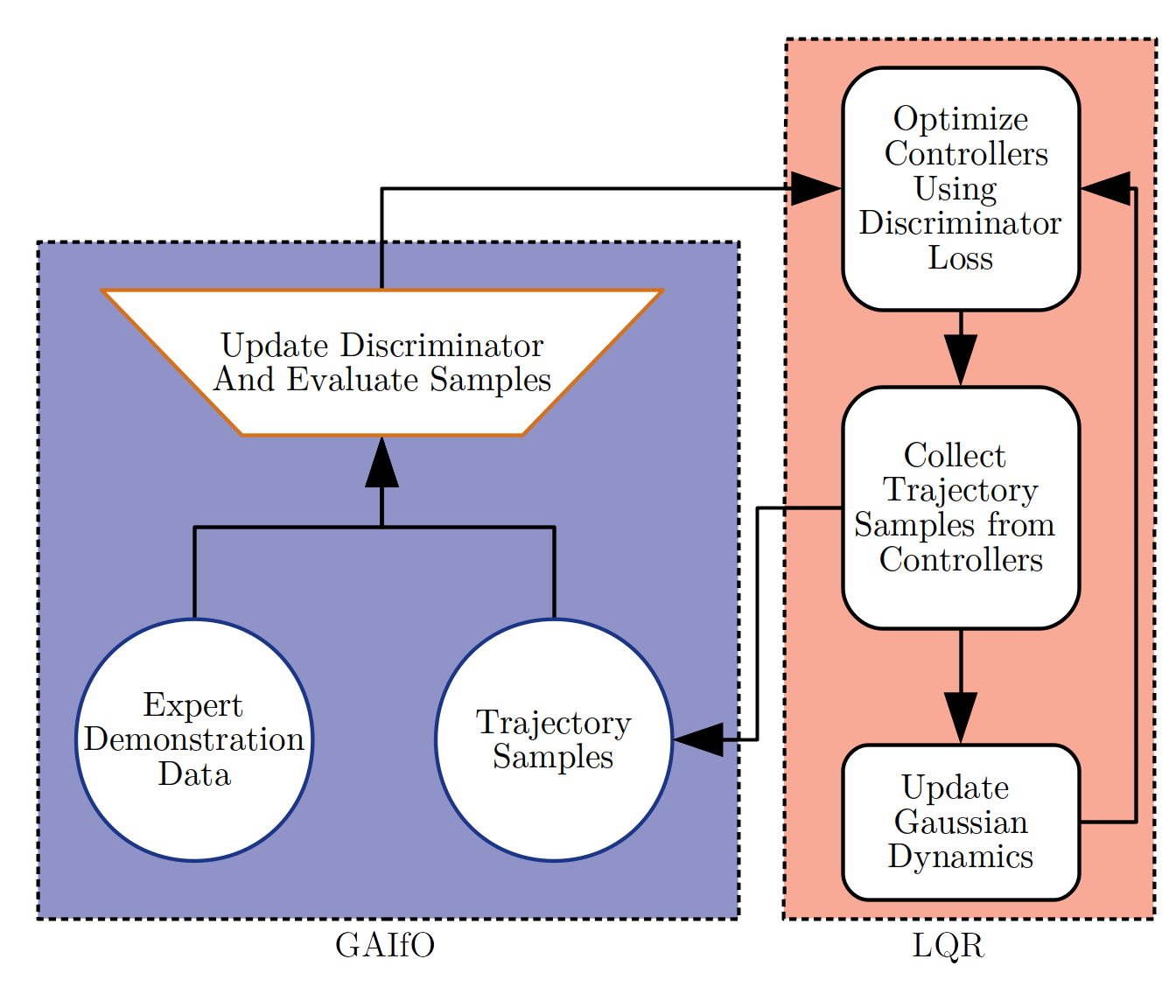}
\par\end{centering}
\caption{The proposed algorithm, LQR+GAIfO.}
\label{fig:algorithm}
\end{figure}

Where $\nabla_{s,a}^{2}$ and $\nabla_{s,a}$ are the Hessian and gradient with respect to the concatenation of $s$ and $a$ vectors, respectively. Finally, an iteration of LQR uses this cost approximation $c_{q}$
to optimize the trajectory to form a new linear-Gaussian controller.
The step size of this update is bounded by the KL-Divergence compared
to the previous iteration. The main components of this approach are
depicted in Figure \ref{fig:algorithm}.

\section{Experiments}\label{sec:exp}

To evaluate the performance of our algorithm, we studied its
ability to imitate a reaching task on a robot arm\textendash both
on a physical arm and in a simulator.

\subsection{Setup}\label{sub:setup}

\begin{figure}
\begin{centering}
\includegraphics[width=0.35\paperwidth]{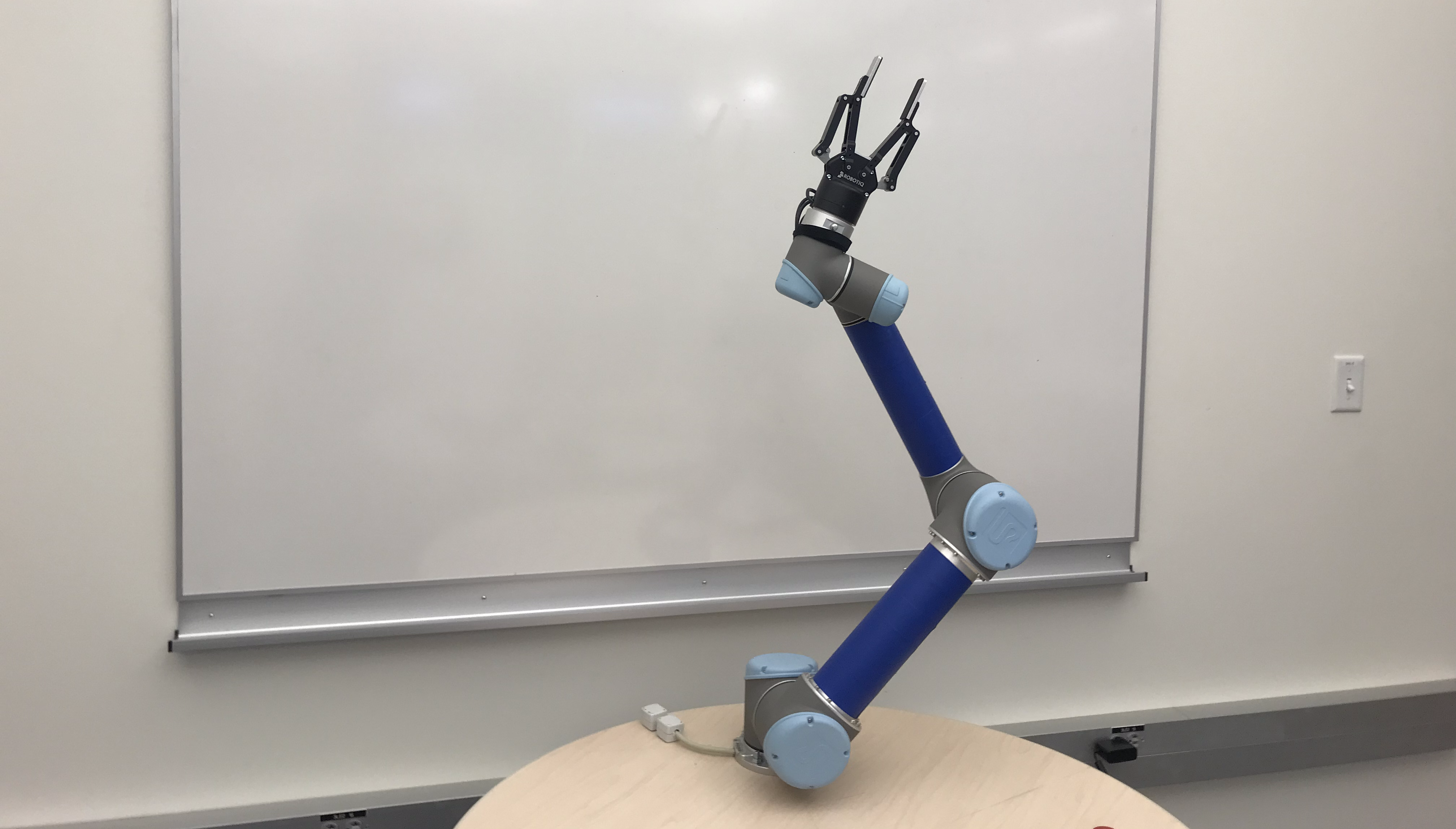}
\par\end{centering}
\caption{The UR5 Robot Arm}
\label{fig:arm}
\end{figure}

\begin{figure}
\begin{centering}
\includegraphics[width=0.42\paperwidth]{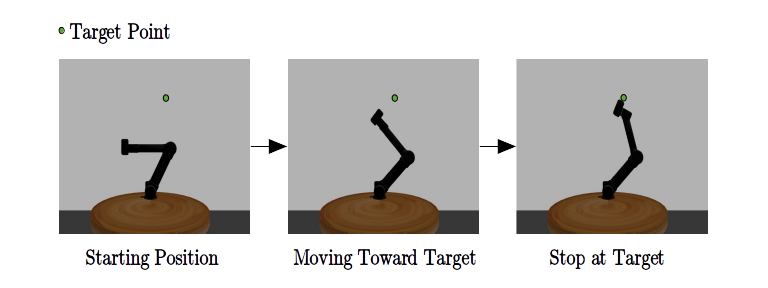}
\par\end{centering}
\caption{A depiction of the reaching task being demonstrated in the simulator.
The arm starts in a retracted position and reaches the end effector
toward the target point, stopping when the target point is reached.}
\label{fig:reaching}
\end{figure}

For a testing platform, we used a Universal Robotics UR5, a 6-degree-of-freedom robotic arm (Figure \ref{fig:arm}). 
%We used an implementation of LQR provided in the GPS implementation by the authors \cite{Levine2016} along with a publically-available integration for the UR5 arm. 
The task that is demonstrated is a reaching task in which the arm begins in a consistent, retracted position and reaches towards a point in Cartesian space. 
When the end effector (the gripper at the end of the arm) reaches this point, the arm stops moving. This task is shown in Figure \ref{fig:reaching}. The expert is trained by iterating between iLQR and dynamics learning with a specified reward function until convergence. 
This policy is then executed and recorded
a number of times to create the demonstration data. 
%For equipment
%safety purposes, only three of the six joints move during both demonstration
%and training, effectively converting the arm to 3-degrees-of-freedom.
%This locks the arm into a single plane of movement (parallel to the
%whiteboard shown in Figure \ref{}, limiting the possibility for collisions
%during training. While leaving joints unused simplifies the state
%space and the dynamics of the arm, we expect the results to generalize
%to higher degrees-of-freedom. 

\begin{figure}
\begin{centering}
\includegraphics[width=0.35\paperwidth]{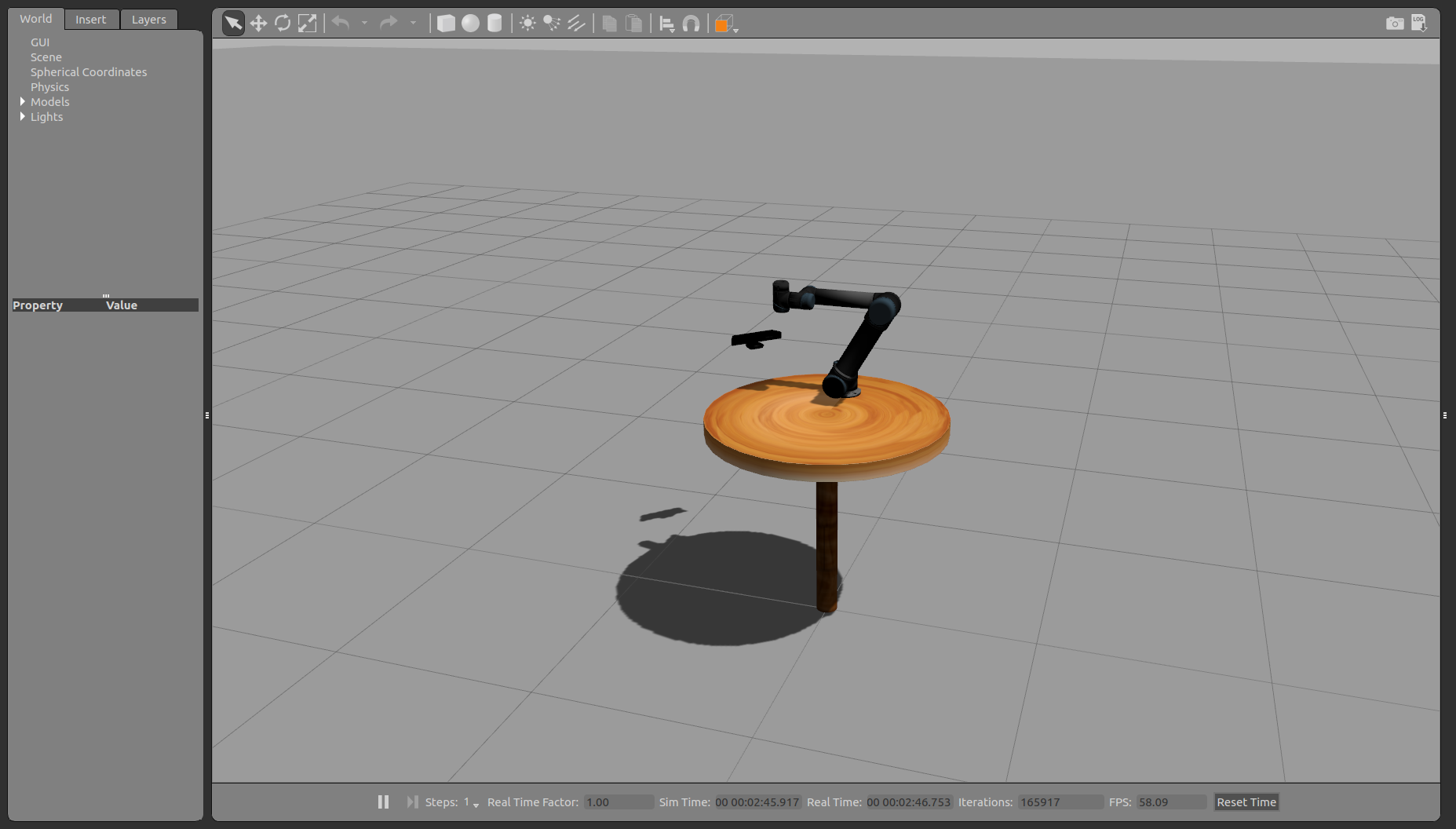}
\par\end{centering}
\caption{The UR5 Arm Modeled in the Gazebo Simulator}
\label{fig:gazebo}
\end{figure}

We modified the software to record the state of the arm and the action
chosen at every time-step of the trajectory execution. For the
initial experiments, the state consisted of:
\begin{enumerate}
\item Joint angles (3 dimensional)
\item Joint velocities (3 dimensional)
\item Cartesian distance to the goal position from the end effector (3 dimensional)
\item Cartesian velocity of the end effector (3 dimensional)
\end{enumerate}
%The action at each time-step (a time-step lasts 0.1 seconds) consists of a vector of goal joint angles for each joint. At each time-step, the
%arm receives a command to move to a goal position. If the arm has
%not reached the goal angles by the next time-step, the arm redirects
%towards the new goal positions. 
For testing in simulation, we used
the Gazebo simulation environment (Figure \ref{fig:gazebo}) with a model of the
UR5. Each trial lasts for 100 timesteps (10 seconds) and ends regardless
of the end effector reaching the goal state. At each iteration, the
policy being evaluated is executed five times to collect five sample
trajectories. The policy is also evaluated once without noise per
iteration, and the performance according to the cost function is logged.

The cost function used takes into account the distance from the end
effector to the target position, weighted linearly as the trial progresses.
With the distance from the goal position to the end effector at a
given time-step $d_{t}$, the cost of a trajectory with $n$ time-steps
is calculated as: 
\[
C(\tau)=d_{t_{n}}+\ensuremath{\sum_{i=0}^{n}\frac{i}{n}d_{t_{i}}}
\]

The same cost function is used to train the expert through reinforcement
learning as well as to evaluate the performance of the imitator. In
this sense, the task of imitation learning can be seen as recovering
the cost function that guided the expert \cite{torabi2018generative}.
For a more complex task or more specific cost function than the one
studied, it's possible that the imitator could recover the task behavior
correctly while not performing well in the eyes of the cost function,
or vice versa. However, for the arm reaching task, the cost function
is simple and directly related to the task, making it appropriate
as an evaluator of imitation performance. For the imitation tasks,
this cost function was used to evaluate each trajectory sample at
a given iteration. The results were normalized on a range from zero
to one, with zero mapping to the average cost of a random policy,
and one mapping to the cost achieved by the expert. A policy
that performs as well as the expert would achieve a score of
one on this normalized performance scale.

We compare our algorithm with GAIfO which is instrumented to interface with the arm control and simulation
platform. Trials for the GAIfO also involved taking
five samples per iteration, in the same way as ours.
The GAIfO policy network was updated using Proximal Policy Optimization (PPO).
%which has been shown to be more sample efficient than the Trust Region
%Policy Optimization (TRPO) \cite{schulman2015trust} approach initially used for GAIL \cite{schulman2017proximal}.

\subsection{Experimental Design}

We conducted three main experiments to evaluate our algorithm. In the
first experiment, the learning rate is compared to learning under
GAIfO. In the second experiment, we test our algorithm's ability
to generalize to unseen target positions. Finally, we compare the performance
of the algorithm in the simulated environment to the physical arm.

\subsubsection{Comparison to GAIfO}

\begin{figure}
\begin{centering}
\includegraphics[width=0.30\paperwidth]{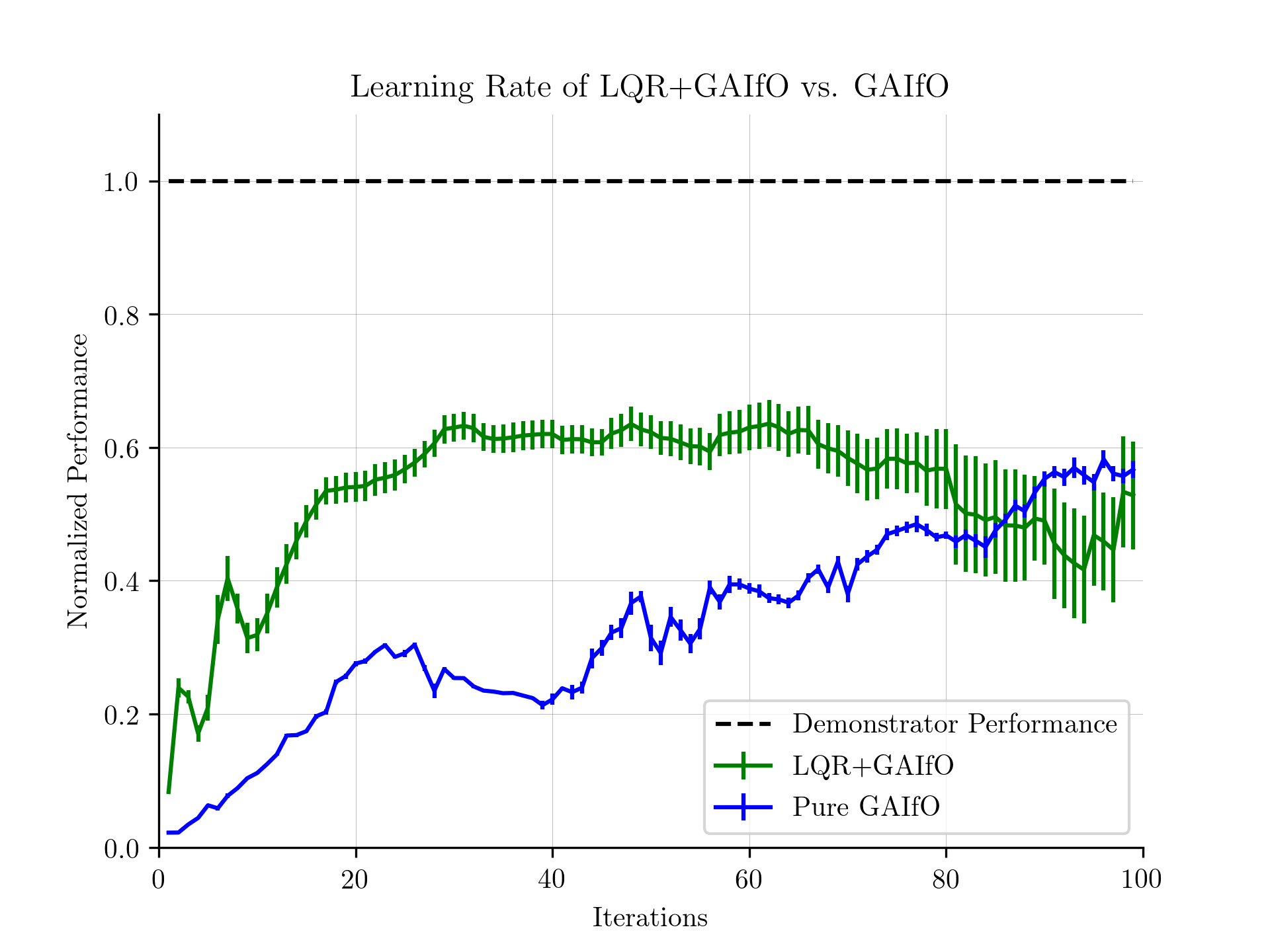}
\par\end{centering}
\caption{Learning rate comparison of LQR+GAIfO to GAIfO in simulation. The
normalized performance is shown, with 0.0 denoting the performance
of a random policy, and 1.0 denoting the performance of the demonstrator.
The error bars show the mean standard error of the policy samples.}
\label{fig:performance}
\end{figure}

To compare the learning rate of our algorithm to that of GAIfO,
we ran trials for both algorithms for 100 iterations and tracked the
policy's performance at each iteration using the cost function described
in Section \ref{sub:setup}. This process was repeated for both algorithms (n=30
for ours, n=55 for GAIfO) to collect average performance data.
The algorithms' performance along with the mean standard error is
plotted in Figure \ref{fig:performance}. The performance of our algorithm quickly exceeds
GAIfO and peaks around iteration 30. 
%Eventually, LQR+GAIfO's
%policy performance starts to degrade around iteration 60 and is passed
%in performance by GAIfO around iteration 90.

\subsubsection{Generalization}

\begin{figure}
\begin{centering}
\includegraphics[width=0.30\paperwidth]{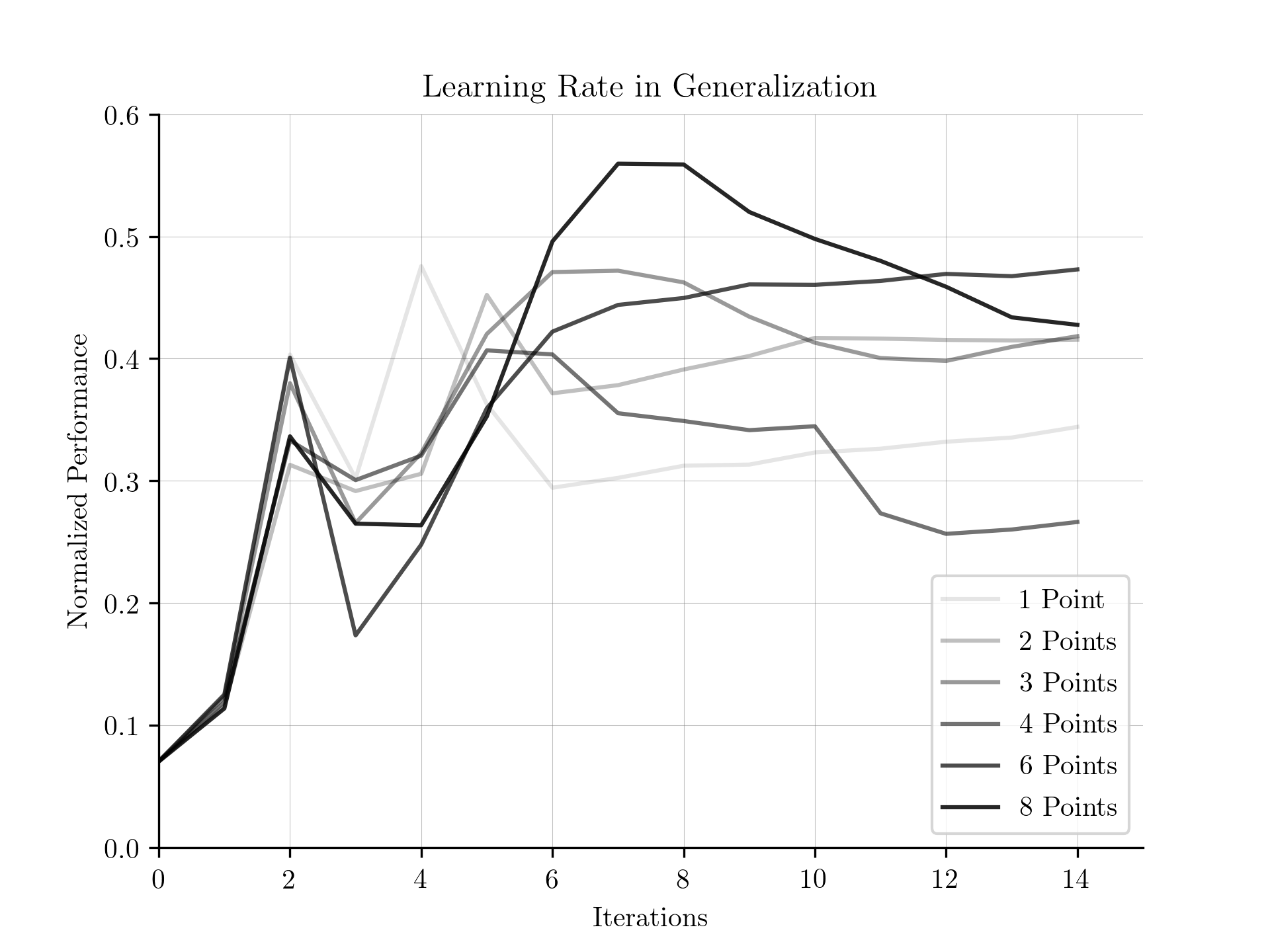}
\par\end{centering}
\caption{The learning rate of LQR+GAIfO in simulation when tasked with reaching an unseen point given a demonstration set with a varying number
of demonstration points.}
\label{fig:multi-point}
\end{figure}

\begin{figure}
\begin{centering}
\includegraphics[width=0.30\paperwidth]{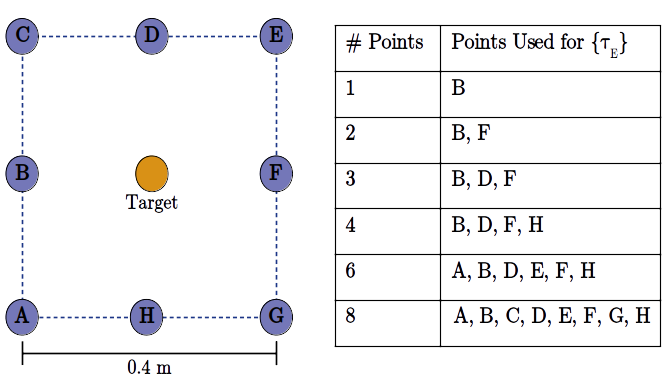}
\par\end{centering}
\centering{}\caption{Points collected for expert data in the generalization experiment.
A varying number of points were chosen from the edges of a square
surrounding the target point. Demonstrated trajectories to these chosen
points form the expert demonstration set.}
\label{fig:points}
\end{figure}

To test our algorithm's ability to generalize a policy for a point that
is not in the expert demonstration data, we collected expert demonstration
trajectories for 8 points on the edge of a square (shown in Figure
\ref{fig:points}). For each point, we trained the expert and recorded five sample
trajectories when the expert converged. Then, after choosing a subset
of the points on the square as $\left\{ \tau_{E}\right\} $, we tasked
the arm with moving to a point in the center of the square. Because
the center point was not in $\left\{ \tau_{E}\right\} $, the control
policy was required to generalize the expert trajectories to this
unseen point. We varied the number of points included in $\left\{ \tau_{E}\right\} $,
and tracked the normalized performance of our algorithm over 15 iterations.
As shown in Figure \ref{fig:multi-point}, while performance was similar in the early iterations,
our algorithm generally performed better in later iterations when more
points were included in $\left\{ \tau_{E}\right\} $.

\subsubsection{Performance on Physical Arm}

\begin{figure}
\begin{centering}
\includegraphics[width=0.30\paperwidth]{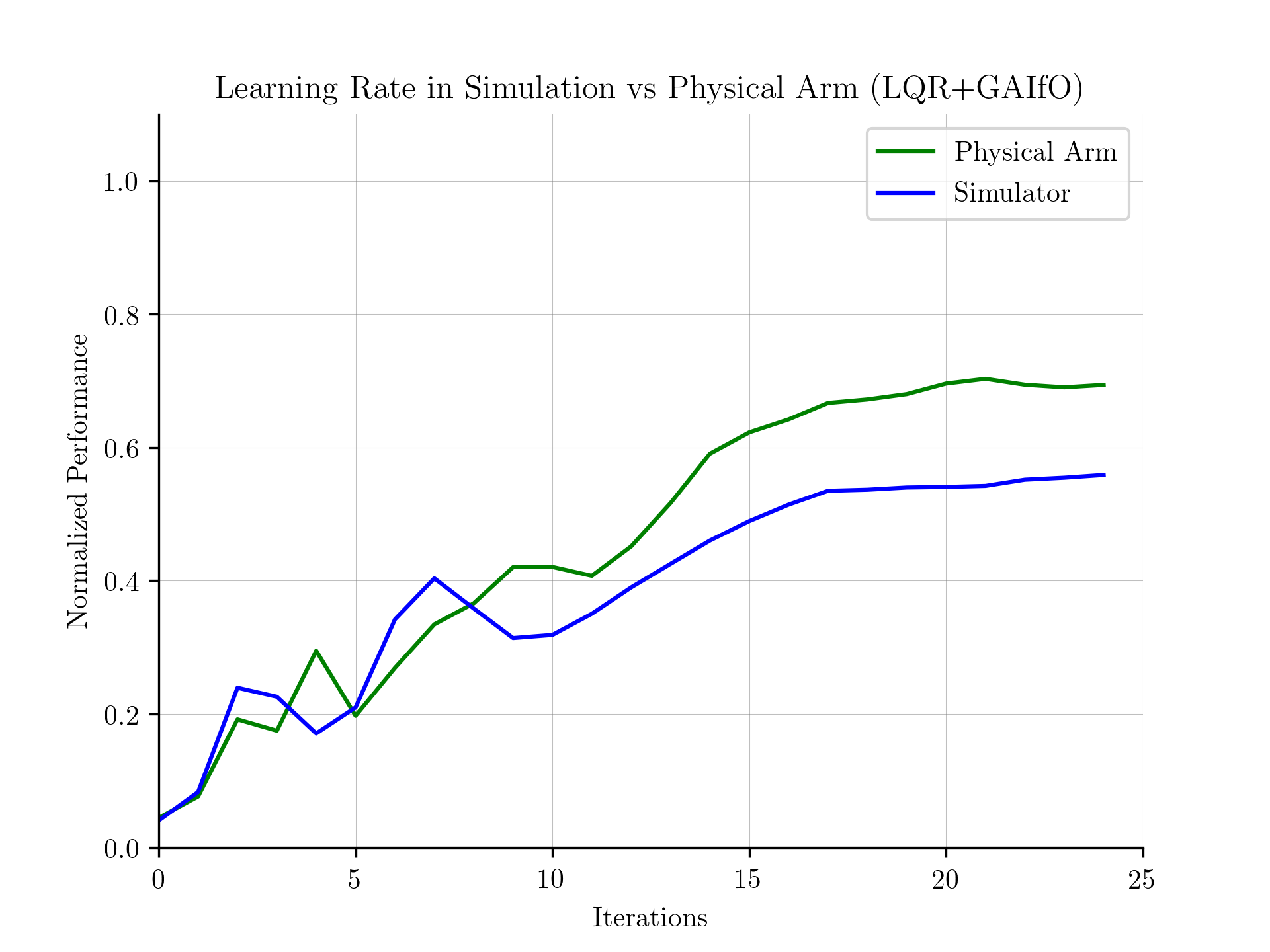}
\par\end{centering}
\caption{The normalized performance of LQR+GAIfO in simulation compared to
the normalized performance of LQR+GAIfO on the physical UR5 arm over
25 iterations.}
\label{fig:comp}
\end{figure}

Our algorithm was run on both the simulator and the physical
arm to examine how closely simulated performance mapped to real-world
performance. Over 25 iterations, the policy performance on the physical
arm began to surpass the performance of the simulated arm, as shown
in Figure \ref{fig:comp}.

\section{Discussion}\label{sec:disc}

Our research began by asking if a combination of LQR and GAIfO could
increase sample efficiency in imitation learning. The comparison of
LQR+GAIfO to GAIfO suggests that LQR+GAIfO can indeed produce a policy
that is better at imitating a behavior in a limited number of iterations,
confirming our hypothesis. The steep initial learning curve of LQR+GAIfO
indicates significantly higher sample efficiency compared to GAIfO
alone. However, the performance of LQR+GAIfO seems to degrade
around iteration 60. Without this performance degradation, LQR+GAIfO
would outperform GAIfO past iteration 100. The reason for this degradation
may be that in adversarial algorithms, improvement of the generator and the discriminator should occur at relatively similar rate. However, in our algorithm, since the controller's representation complexity is limited, after some number of iteration, the controller does not improve as fast as the discriminator.
In addition, even without this degradation, the GAIfO approach would
eventually surpass the performance of LQR+GAIfO, likely due to the
ability of the generator network in GAIfO to produce more complex
policies than those that can be represented with linear Gaussian controllers
in LQR.

Although most of the ability for a policy to perform a task that is
different from the expert trajectories in GAIfO and GPS result from
a complex model considered for the policy (neural network), the linear Gaussian controllers in LQR+GAIfO still have
the ability to generalize to some degree. As expected, the ability
to successfully generalize increases with demonstration trajectories,
as shown in Figure \ref{fig:multi-point}. The reason may be that the discriminator learns a general cost function that could be applied to new target points and as a result LQR can learn a relatively good controller. Future work integrating the
full GPS approach would likely lead to better generalization.

We studied the performance of LQR+GAIfO on the physical arm to validate
the tractability of this technique on a real robot and to establish
a sense of how directly the performance studied in the simulator would
translate to the physical arm. Our results, as seen in Figure \ref{fig:comp}, show
that the policy performance seen in the simulator can be trusted to
model policy performance on the real arm. Surprisingly, the performance
of LQR+GAIfO on the physical arm exceeds the simulator performance. It is possible that the
noise introduced by the physical arm as a result of actuator noise
or other physical effects lead to wider exploration and faster policy
improvement. If this is the case, it could be possible to achieve
similar performance in the simulator by introducing more policy noise.
%Further experimentation is required to fully understand this effect.

\section{Conclusion and Future Work}\label{sec:conclusion}

We have found that combining generative adversarial imitation from
observation with Linear Quadratic Regulators leads to faster learning
of imitation behavior over fewer samples than with GAIfO alone, confirming
our hypothesis. While LQR+GAIfO doesn't reach the absolute imitation
performance of GAIfO over an extended training period with thousands
of samples, achieving adequate imitation performance with limited
samples opens the door to imitation research on physical robotic systems,
for which imitation learning has posed logistical challenges in the
past.

While LQR is a powerful technique by itself, a policy based solely
on Gaussian controllers has limits in complexity. Work in GPS has
already produced a method for combining sample-efficient Gaussian
controllers with a deep network model that is trained through the
controllers. Using a deep network as part of the policy offers increased
performance in the long run and greatly increased generalization ability.
Incorporating this deep network policy driven by importance-weighted
samples of the linear Gaussian controllers is an obvious and promising
next step for this work.

To validate the LQR+GAIfO technique, we represented the expert trajectories
using low-level data like the Cartesian position of the arm's end
effector. GAIfO has had success in using higher level data\textendash like
a visual recording of the demonstrator\textendash as the state in
trajectories. Additionally, GPS has been used in learning neural network policies from visual observation \cite{levine2016end}.
Pursuing imitation learning from visual data alone would greatly widen
the situations in which demonstration data could be collected. Adding
a convolutional layer to the discriminator so that it can accept visual
data is a natural next step for extending this research.

\section*{Acknowledgements}
This work has taken place in the Learning Agents Research
Group (LARG) at the Artificial Intelligence Laboratory, The University
of Texas at Austin.  LARG research is supported in part by grants from
the National Science Foundation (IIS-1637736, IIS-1651089,
IIS-1724157), the Office of Naval Research (N00014-18-2243), Future of
Life Institute (RFP2-000), Army Research Lab, DARPA, Intel, Raytheon,
and Lockheed Martin.  Peter Stone serves on the Board of Directors of
Cogitai, Inc.  The terms of this arrangement have been reviewed and
approved by the University of Texas at Austin in accordance with its
policy on objectivity in research.


\bibliography{icml2019}

\begin{thebibliography}{31}
\providecommand{\natexlab}[1]{#1}
\providecommand{\url}[1]{\texttt{#1}}
\expandafter\ifx\csname urlstyle\endcsname\relax
  \providecommand{\doi}[1]{doi: #1}\else
  \providecommand{\doi}{doi: \begingroup \urlstyle{rm}\Url}\fi

\bibitem[Argall et~al.(2009)Argall, Chernova, Veloso, and Browning]{Argall2009}
Argall, B.~D., Chernova, S., Veloso, M., and Browning, B.
\newblock A survey of robot learning from demonstration.
\newblock \emph{Robotics and autonomous systems}, 57\penalty0 (5):\penalty0
  469--483, 2009.

\bibitem[Arjovsky et~al.(2017)Arjovsky, Chintala, and
  Bottou]{arjovsky2017wasserstein}
Arjovsky, M., Chintala, S., and Bottou, L.
\newblock Wasserstein generative adversarial networks.
\newblock In \emph{International Conference on Machine Learning}, pp.\
  214--223, 2017.

\bibitem[Bemporad et~al.(2002)Bemporad, Morari, Dua, and
  Pistikopoulos]{Bemporad2002}
Bemporad, A., Morari, M., Dua, V., and Pistikopoulos, E.~N.
\newblock {The explicit linear quadratic regulator for constrained systems}.
\newblock \emph{Automatica}, 38\penalty0 (1):\penalty0 3--20, 2002.
\newblock ISSN 00051098.
\newblock \doi{10.1016/S0005-1098(01)00174-1}.

\bibitem[Chen et~al.(2016)Chen, Duan, Houthooft, Schulman, Sutskever, and
  Abbeel]{chen2016infogan}
Chen, X., Duan, Y., Houthooft, R., Schulman, J., Sutskever, I., and Abbeel, P.
\newblock Infogan: Interpretable representation learning by information
  maximizing generative adversarial nets.
\newblock In \emph{Advances in neural information processing systems}, pp.\
  2172--2180, 2016.

\bibitem[Finn et~al.(2016)Finn, Levine, and Abbeel]{finn2016guided}
Finn, C., Levine, S., and Abbeel, P.
\newblock Guided cost learning: Deep inverse optimal control via policy
  optimization.
\newblock In \emph{International Conference on Machine Learning}, pp.\  49--58,
  2016.

\bibitem[Fu et~al.(2018)Fu, Luo, and Levine]{fu2018learning}
Fu, J., Luo, K., and Levine, S.
\newblock Learning robust rewards with adversarial inverse reinforcement
  learning.
\newblock 2018.

\bibitem[Goodfellow et~al.(2014)Goodfellow, Pouget-Abadie, Mirza, Xu,
  Warde-Farley, Ozair, Courville, and Bengio]{goodfellow2014generative}
Goodfellow, I., Pouget-Abadie, J., Mirza, M., Xu, B., Warde-Farley, D., Ozair,
  S., Courville, A., and Bengio, Y.
\newblock Generative adversarial nets.
\newblock In \emph{Advances in neural information processing systems}, pp.\
  2672--2680, 2014.

\bibitem[Gulrajani et~al.(2017)Gulrajani, Ahmed, Arjovsky, Dumoulin, and
  Courville]{gulrajani2017improved}
Gulrajani, I., Ahmed, F., Arjovsky, M., Dumoulin, V., and Courville, A.~C.
\newblock Improved training of wasserstein gans.
\newblock In \emph{Advances in Neural Information Processing Systems}, pp.\
  5767--5777, 2017.

\bibitem[Ho \& Ermon(2016)Ho and Ermon]{ho2016generative}
Ho, J. and Ermon, S.
\newblock Generative adversarial imitation learning.
\newblock In \emph{Advances in Neural Information Processing Systems}, pp.\
  4565--4573, 2016.

\bibitem[Ijspeert et~al.(2001)Ijspeert, Nakanishi, and
  Schaal]{ijspeert2001trajectory}
Ijspeert, A.~J., Nakanishi, J., and Schaal, S.
\newblock Trajectory formation for imitation with nonlinear dynamical systems.
\newblock In \emph{Proceedings 2001 IEEE/RSJ International Conference on
  Intelligent Robots and Systems. Expanding the Societal Role of Robotics in
  the the Next Millennium (Cat. No. 01CH37180)}, volume~2, pp.\  752--757.
  IEEE, 2001.

\bibitem[Kostrikov et~al.(2019)Kostrikov, Agrawal, Dwibedi, Levine, and
  Tompson]{kostrikov2019addressing}
Kostrikov, I., Agrawal, K.~K., Dwibedi, D., Levine, S., and Tompson, J.
\newblock Discriminator-actor-critic: Addressing sample inefficiency and reward
  bias in adversarial imitation learning.
\newblock 2019.

\bibitem[Levine \& Abbeel(2014)Levine and Abbeel]{levine2014learning}
Levine, S. and Abbeel, P.
\newblock Learning neural network policies with guided policy search under
  unknown dynamics.
\newblock In \emph{Advances in Neural Information Processing Systems}, pp.\
  1071--1079, 2014.

\bibitem[Levine \& Koltun(2013)Levine and Koltun]{Levine2013}
Levine, S. and Koltun, V.
\newblock {Guided Policy Search}.
\newblock \emph{Proceedings of the 30th International Conference on Machine
  Learning}, 28:\penalty0 1--9, 2013.
\newblock URL \url{http://jmlr.org/proceedings/papers/v28/levine13.html}.

\bibitem[Levine et~al.(2015)Levine, Wagener, and Abbeel]{Levine2015a}
Levine, S., Wagener, N., and Abbeel, P.
\newblock {Learning contact-rich manipulation skills with guided policy
  search}.
\newblock \emph{Proceedings - IEEE International Conference on Robotics and
  Automation}, 2015-June\penalty0 (June):\penalty0 156--163, 2015.
\newblock ISSN 10504729.
\newblock \doi{10.1109/ICRA.2015.7138994}.

\bibitem[Levine et~al.(2016)Levine, Finn, Darrell, and Abbeel]{levine2016end}
Levine, S., Finn, C., Darrell, T., and Abbeel, P.
\newblock {End-to-end training of deep visuomotor policies}.
\newblock \emph{The Journal of Machine Learning Research}, 17\penalty0
  (1):\penalty0 1334--1373, 2016.

\bibitem[Li \& Todorov(2004)Li and Todorov]{li2004iterative}
Li, W. and Todorov, E.
\newblock {Iterative linear quadratic regulator design for nonlinear biological
  movement systems.}
\newblock In \emph{ICINCO (1)}, pp.\  222--229, 2004.

\bibitem[Liu et~al.(2018)Liu, Gupta, Abbeel, and Levine]{liu2018imitation}
Liu, Y., Gupta, A., Abbeel, P., and Levine, S.
\newblock Imitation from observation: Learning to imitate behaviors from raw
  video via context translation.
\newblock In \emph{2018 IEEE International Conference on Robotics and
  Automation (ICRA)}, pp.\  1118--1125. IEEE, 2018.

\bibitem[Ng et~al.(2000)Ng, Russell, et~al.]{ng2000algorithms}
Ng, A.~Y., Russell, S.~J., et~al.
\newblock Algorithms for inverse reinforcement learning.
\newblock In \emph{Icml}, volume~1, pp.\ ~2, 2000.

\bibitem[Pomerleau(1991)]{Pomerleau1991}
Pomerleau, D.
\newblock {Efficient Training of Artificial Neural Networks for Autonomous
  Navigation}.
\newblock \emph{Neural Computation}, 1991.

\bibitem[Russell(1998)]{russell1998learning}
Russell, S.~J.
\newblock {Learning agents for uncertain environments}.
\newblock In \emph{COLT}, volume~98, pp.\  101--103, 1998.

\bibitem[Schulman et~al.(2015)Schulman, Levine, Abbeel, Jordan, and
  Moritz]{schulman2015trust}
Schulman, J., Levine, S., Abbeel, P., Jordan, M., and Moritz, P.
\newblock Trust region policy optimization.
\newblock In \emph{International Conference on Machine Learning}, pp.\
  1889--1897, 2015.

\bibitem[Song et~al.(2018)Song, Ren, Sadigh, and Ermon]{song2018multi}
Song, J., Ren, H., Sadigh, D., and Ermon, S.
\newblock Multi-agent generative adversarial imitation learning.
\newblock In \emph{Advances in Neural Information Processing Systems}, pp.\
  7461--7472, 2018.

\bibitem[Stadie et~al.(2017)Stadie, Abbeel, and Sutskever]{stadie2017third}
Stadie, B.~C., Abbeel, P., and Sutskever, I.
\newblock Third-person imitation learning.
\newblock 2017.

\bibitem[Sutton \& Barto(1998)Sutton and Barto]{Sutton1998}
Sutton, R. and Barto, A.
\newblock \emph{{Reinforcement Learning: An introduction}}.
\newblock MIT Press, Cambridge, volume 1 edition, 1998.

\bibitem[Tassa et~al.(2012)Tassa, Erez, and Todorov]{tassa2012synthesis}
Tassa, Y., Erez, T., and Todorov, E.
\newblock Synthesis and stabilization of complex behaviors through online
  trajectory optimization.
\newblock In \emph{2012 IEEE/RSJ International Conference on Intelligent Robots
  and Systems}, pp.\  4906--4913. IEEE, 2012.

\bibitem[Torabi et~al.(2018{\natexlab{a}})Torabi, Warnell, and
  Stone]{torabi2018behavioral}
Torabi, F., Warnell, G., and Stone, P.
\newblock Behavioral cloning from observation.
\newblock In \emph{Proceedings of the 27th International Joint Conference on
  Artificial Intelligence}, pp.\  4950--4957. AAAI Press, 2018{\natexlab{a}}.

\bibitem[Torabi et~al.(2018{\natexlab{b}})Torabi, Warnell, and
  Stone]{torabi2018generative}
Torabi, F., Warnell, G., and Stone, P.
\newblock {Generative adversarial imitation from observation}.
\newblock \emph{arXiv preprint arXiv:1807.06158}, 2018{\natexlab{b}}.

\bibitem[Torabi et~al.(2019{\natexlab{a}})Torabi, Warnell, and
  Stone]{torabi2019adversarial}
Torabi, F., Warnell, G., and Stone, P.
\newblock Adversarial imitation learning from state-only demonstrations.
\newblock In \emph{International Conference on Autonomous Agents and
  Multi-Agent Systems}, 2019{\natexlab{a}}.

\bibitem[Torabi et~al.(2019{\natexlab{b}})Torabi, Warnell, and
  Stone]{torabi2019imitation}
Torabi, F., Warnell, G., and Stone, P.
\newblock Imitation learning from video by leveraging proprioception.
\newblock In \emph{International Joint Conference on Artificial Intelligence},
  2019{\natexlab{b}}.

\bibitem[Torabi et~al.(2019{\natexlab{c}})Torabi, Warnell, and
  Stone]{torabi2019recent}
Torabi, F., Warnell, G., and Stone, P.
\newblock Recent advances in imitation learning from observation.
\newblock In \emph{International Joint Conference on Artificial Intelligence}.
  AAAI Press, 2019{\natexlab{c}}.

\bibitem[Ziebart et~al.(2008)Ziebart, Maas, Bagnell, and
  Dey]{ziebart2008maximum}
Ziebart, B.~D., Maas, A.~L., Bagnell, J.~A., and Dey, A.~K.
\newblock Maximum entropy inverse reinforcement learning.
\newblock In \emph{AAAI}, volume~8, pp.\  1433--1438. Chicago, IL, USA, 2008.

\end{thebibliography}
\bibliographystyle{icml2019}

\end{document}